\documentclass[letterpaper, 10 pt, conference]{ieeeconf}  

\IEEEoverridecommandlockouts                              

\usepackage{graphics} 
\usepackage{epsfig} 
\usepackage{amsmath} 
\usepackage{amssymb}  
\usepackage{bm} 
\usepackage[flushmargin,hang]{footmisc} 
\usepackage{flushend} 
\usepackage[acronym]{glossaries}

\def\epsgaiji#1{\leavevmode\kern-0.025zw\raise-.37zh\hbox{%
  \epsfile{file=#1,width=1.05zw}}\kern-0.025zw}
\newcommand{\MARU}[1]{{\ooalign{\hfil#1\/\hfil\crcr\raise.167ex\hbox{\mathhexbox20D}}}}

\usepackage{array}

\newacronym{lro}    {LRO}   {Lunar Reconnaissance Orbiter}
\newacronym{usd}    {USD}   {Universal Scene Description}
\newacronym{slam}    {SLAM}   {Simulataneous Localization And Mapping}
\newacronym{lroc}   {LROC}  {Lunar Reconnaissance Orbiter Camera}
\newacronym{lola}   {LOLA}  {Lunar Orbiter Laser Altimeter}
\newacronym{dem}    {DEM}   {Digital Elevation Map}
\newacronym{nac}    {NAC}   {Narrow Angle Camera}
\newacronym{nn}     {NN}    {Neural Network}
\newacronym{cnn}    {CNN}   {Convolutional Neural Network}
\newacronym{gan}    {GAN}   {Generative Adversarial Network}
\newacronym{sr}     {SR}    {Super Resolution}
\newacronym{asp}    {ASP}   {Ames Stereo Pipelines}
\newacronym{sfs}    {SFS}   {Shape-From-Shading}
\newacronym{wac}    {WAC}   {Wide Angle Camera}
\newacronym{lr}     {LR}    {Low-Resolution}
\newacronym{hr}     {HR}    {High-Resolution}
\newacronym{mpp}    {mpp}   {meters per pixel}
\newacronym{mse}    {MSE}   {Mean Square Error}
\newacronym{rmse}   {RMSE}  {Root Mean Square Error}
\newacronym{mae}   {MAE}  {Mean Average Error}
\newacronym{lod}    {LoD}   {Level of Detail}
\newacronym{ue}    {UE}   {Unreal Engine}

\usepackage{siunitx} 
\usepackage{gensymb} 
\usepackage{multirow} 

\usepackage{hyperref}
\hypersetup{
    colorlinks=true,
    linkcolor=blue,
    filecolor=magenta,      
    urlcolor=cyan,
    pdftitle={Overleaf Example},
    pdfpagemode=FullScreen,
    }
\hypersetup{hidelinks}

\usepackage{pgfplots}
\pgfplotsset{compat=newest}
\usetikzlibrary{plotmarks}
\usetikzlibrary{arrows.meta}
\usepgfplotslibrary{patchplots}
\usepackage{grffile}
\pgfplotsset{plot coordinates/math parser=false}
\newlength\fwidth
\newlength\fheight

\usepackage{cite}
\usepackage{subfig}
\usepackage{url}

\usepackage{pifont}
\usepackage{amssymb}

\title{\LARGE \bf
OmniLRS: A Photorealistic Simulator for Lunar Robotics}

\author{
Antoine Richard$^{1}$*, Junnosuke Kamohara$^{2}$*, Kentaro Uno$^{2}$, Shreya Santra$^{2}$, Dave van der Meer$^{1}$\\ Miguel Olivares-Mendez$^{1}$ and Kazuya Yoshida$^{2}$
\thanks{$^{1}$ A. Richard, D. van der Meer, M. Olivares-Mendez are with The Space Robotics Research Group at the Interdisciplinary Research Center for Security reliability and Trust (SnT) in the University of Luxembourg {\tt\small antoine.richard@uni.lu}}%
\thanks{$^{2}$ J. Kamohara, K. Uno, S. Santra and K. Yoshida are with the Space Robotics Lab. in Department of Aerospace
Engineering, Graduate School of Engineering, Tohoku University, Sendai 980–8579, Japan.
        {\tt\small kamohara.junnosuke.t6@dc.tohoku.ac.jp}}%
\thanks{*These authors contributed equally.}
}%

\begin{document}

\maketitle
\thispagestyle{empty}
\pagestyle{empty}


\begin{abstract}
Developing algorithms for extra-terrestrial robotic exploration has always been challenging.
Along with the complexity associated with these environments, one of the main issues remains the evaluation of said algorithms.
With the regained interest in lunar exploration, there is also a demand for quality simulators that will enable the development of lunar robots.
In this paper, we propose Omniverse Lunar Robotic-Sim (OmniLRS) that is a photorealistic Lunar simulator based on Nvidia's robotic simulator.
This simulation provides fast procedural environment generation, multi-robot capabilities, along with synthetic data pipeline for machine-learning applications.
It comes with ROS1 and ROS2 bindings to control not only the robots, but also the environments.
This work also performs sim-to-real rock instance segmentation to show the effectiveness of our simulator for image-based perception. 
Trained on our synthetic data, a yolov8 model achieves performance close to a model trained on real-world data, with 5\% performance gap.
When finetuned with real data, the model achieves 14\% higher average precision than the model trained on real-world data, demonstrating our simulator's photorealism.
The code is fully open-source, accessible here: \url{https://github.com/AntoineRichard/LunarSim}, and comes with demonstrations.
\end{abstract}


\section{INTRODUCTION}
There has never been as much interest in sending robots to the Moon as today.
With this renewed interest, carried by both institutional and private entities, comes a need to test the software that will enable the robots to carry out their missions.
Typically, this is done in two stages, the first stage in a simulator, and the second stage in a lunar analog environment, such as Mount Etna.
In the first stage, a wide range of simulators are employed to be able to assess all the systems required to complete the mission.
These simulators range from dedicated simulation of mechanical parts, through landing, to robotics.
Within robotics itself, there are often multiple sub-simulators, such as for terramechanics interactions, for thermal resilience of the system, for perception data, and many more.
In this study, we focus on providing an accurate visual representation of the lunar environment with typical perception sensors, such as RGB cameras, depth imaging, or Lidars.
This kind of simulator is well suited for vision-centric tasks such as navigation, exploration, \gls{slam}, and machine vision.

A common issue with this type of simulator is that more often than not, they are not available to the general public~\cite{NASA-DLES, DUST, ViperGazeboSim, Pangu}, or paid for~\cite{Pangu}.
This is often true with the simulators developed by private companies or institutional entities, such as NASA or ESA.
Others, can have some strong limitations with regard to rendering capabilities~\cite{Gazebo, orsula2022learning, kilic2023multi}, or flexibility~\cite{FauxRanger, PUT-sim, edge}.
To the best of our knowledge, there is a lack of easy-to-use, flexible robotics lunar simulators, that also provide good render quality.
Most of the time, to achieve high-quality renders, developers have built upon commercial game engines such as \gls{ue}~\cite{DUST, PUT-sim, FauxRanger} or Unity~\cite{NASA-DLES, edge}.
Both of these can be daunting to modify to accommodate robotics systems and their sensors, as they require advanced coding skills and substantial development efforts.
This also hinders the flexibility of the simulator built upon these engines: adding new sensors or robots can be challenging, even more so if they do not adopt standard formats.
Another issue that often arises with most of these simulators are the algorithms used to build the terrains.
These algorithms, such as \gls{asp}~\cite{AmesSP} can be very complex to use, and compute-intensive. 


\begin{figure}[t]
    \centering
    \includegraphics[width=85mm]{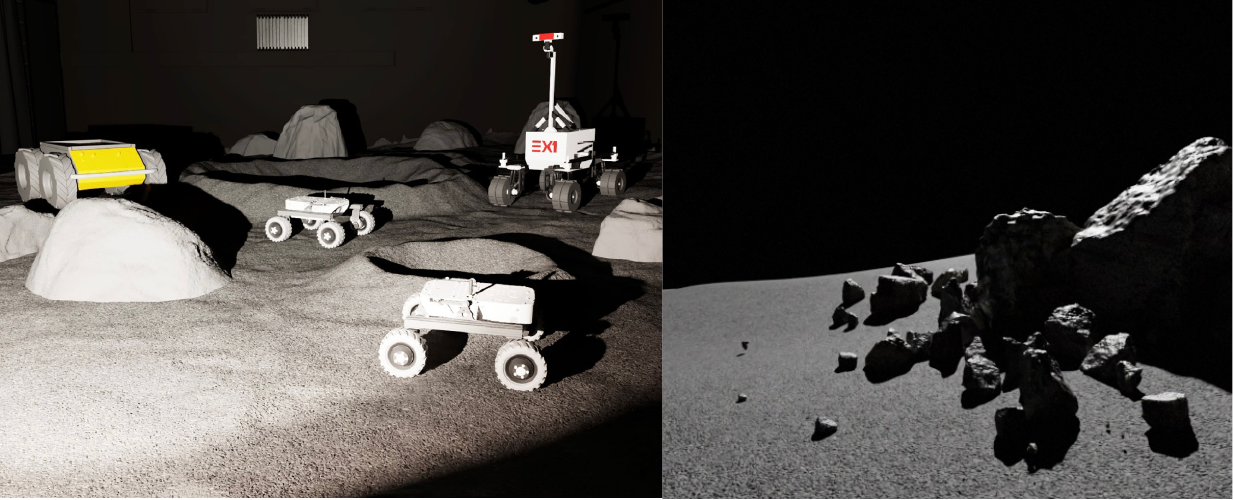}
    \caption{
        A render of Lunalab (left) and a real lunar terrain (right) using pathtraced rendering with OmniLRS.
        The environment includes three types of rover models: EX1, Leo rover, and Husky.
    }
    \label{fig:cover-img}
    \vspace{-5mm}
\end{figure}

With this paper, we chose to build upon IsaacSim, a so-called application of Nvidia's Omniverse.
Similarly to Gazebo~\cite{Gazebo}, IsaacSim provides built-in robotics sensors so that the users do not have to implement them, and it offers compatibility layers for generic robotic formats such as URDF and SDF.
This means that, like Gazebo-based solutions~\cite{ViperGazeboSim}, our simulator offers strong flexibility, and should be easy to modify.
However, unlike Gazebo, IsaacSim comes with a powerful render engine that can use either ray-traced or path-traced renderers.
Fig.~\ref{fig:cover-img} shows a sample render of the simulated Lunalab using path-tracing.
This is particularly important for environments such as the moon, where correct illumination plays a key role in the visual appearance of the scene.
Furthermore, since Isaac builds upon the \gls{usd}~\cite{PixarUSD}, it can use instanced assets and variants, allowing it to render millions of instanced assets in real time on a small GPU.
Something that, to our knowledge, isn't possible in Gazebo.
Overall, with this simulation, we provide a set of open-source tools enabling anyone to easily simulate lunar environments, real or fake, and perform robotic activities on them.
We also provide means to collect datasets for machine vision tasks, such as instance segmentation or object detection.
Finally, we provide a set of demonstrations in ROS1 and ROS2, showing how to run navigation, mapping, \gls{slam}, or object detection.

In addition to the implementation of the lunar simulator, we evaluate the realism of the generated environment by training a rock instance segmentation model~\cite{yolov8}.
With the model trained on different sets of data, both real and synthetic, we investigate the following: the feasibility of zero-shot sim-to-real, and the possibility of leveraging synthetic data as a base model that can be improved by finetuning with real-world data.

In summary, our contributions are the following: 1) A ROS enabled simulation framework for lunar environments based on IsaacSim, fully open source, with readily available terrains, assets, and robots; 2) A sim-to-real study of rock instance segmentation in a lunar analog environment. 



\section{Related Work}
\subsection{Lunar Simulators}
As of today, there is a diversity of lunar ground simulators developed with different graphics engines. 
All of these aim at testing software components, including but not limited to: navigation, manipulation, SLAM, or even full-scale mission planning, assembly and resource exploration.

DLES (Digital Lunar Exploration Sites)~\cite{NASA-DLES} leverages LRO and LOLA \glspl{dem} to accurately model the lunar South Pole.
In addition to terrain generation, it also adds features such as small craters and rocks of different sizes which cannot be captured in \gls{dem} to emulate the lunar surface in a more realistic way.
By upscaling five meters per pixel LOLA \gls{dem} and overlaying craters, DLES achieves 20 centimeters per pixel \gls{dem}.
High-resolution terrain models along with surface details, rocks and craters, from DLES are incorporated into DLES Unreal Simulation Tool (DUST)~\cite{DUST} to simulate a photorealistic lunar South Pole.
It leverages the advanced capabilities of \gls{ue}5, such as Lumen to represent Sun and Earth-shine, virtual shadow maps~\cite{Virtual-Shadow-Map} to display detailed shadows over thousands of kilometers of terrain, real-time ray-tracing, and Nanite to display large detailed meshes efficiently.
With DUST, the DLES generated maps are rendered with impressive fidelity.
Yet, unfortunately, DUST does not support robotic systems and is not available to the public.
\cite{orsula2022learning} uses Gazebo, to train and infer RL policies to grab lunar-rocks, this simulator is mostly meant for grasping, and as such, features small terrains non-applicable for other vision tasks.
\cite{PUT-sim, FauxRanger} are both open-source and use \gls{ue}4 to simulate the environment.
They are meant for navigation tasks and provide good rendering capabilities.
Yet, they do not support the URDF format, and do not offer conversion tools to take an existing Gazebo-ready robot and integrate it in their environments.
This is also true of~\cite{edge} which builds on Unity.
Overall, this hinders their use, as changing the robot or adding sensors is non-trivial.
Another issue is that both ~\cite{PUT-sim, FauxRanger} are building on top of a discontinued engine, and the process they are using to generate or load terrains is unclear.

The simulator used in the Viper mission~\cite{ViperGazeboSim} based on gazebo, provides a lot of must have-features for lunar explorations, such as the management of large terrain with fine details, the generation of lens-flares, realistic shadows, as well as the opposition effect.
Yet, to achieve this, they heavily modified the render engine from gazebo, and added some novel features which did not make their way to the main branch of gazebo.
Moreover, this simulator is currently a ``NASA internal only tool'' making it inaccessible to the public.

\subsection{Dataset from Extraterrestrial Environment}
Synthetic data generation is now a typical part of any machine vision tasks when working with limited real-data~\cite{metasim, metasim2}.
In this subsection, we provide an overview of existing dataset for vision-based recognition of rocks, environmental landmarks that could help \gls{slam} algorithms in low feature environments like Moon and Mars. 

\indent
Artificial Lunar Landscape Dataset~\cite{kaggle-lunar-dataset} is a synthetic lunar landscape dataset for rock detection and semantic segmentation.
The landscape is created by Terragen using LRO LOLA \glspl{dem} recorded at the lunar South Pole, with added fractal noise and rocks.
The data consists of 9,766 synthetic renderings, and also includes 22 annotated images from Chang'e3 mission.

ReSyRIS (Real-Synthetic Rock Instance Segmentation Dataset)~\cite{ReSyRIS} is a collection of real-world images from lunar analog environments and a synthetic counterpart for rock instance segmentation.
The real-world data is collected in Mt. Etna, and annotated with pixel-wise instance masks.
For the synthetic data, they use OAISYS~\cite{DLR-OAISYS}, a Blender-based simulator of outdoor unstructured environments featuring automatic annotations.
These data are then used to investigate sim-to-real gap.

While both~\cite{ReSyRIS} and~\cite{kaggle-lunar-dataset} offer very realistic renders, they are fixed datasets, which means that novel data cannot be generated.

Aside from lunar images, there also exist publicly available Martian data~\cite{GMSRI,Zhang2022S5MarsSA, AI4Mars}. 
AI4Mars~\cite{AI4Mars} is a public large-scale dataset for training and validating terrain classification models such as Soil Property and Object Classification (SPOC)~\cite{SPOCDL}.
The dataset contains 35,000 images from Curiosity, Opportunity, and Spirit rovers.
It focuses on the traversability assessment on Mars, and provides 4 semantic classes (soil, bedrock, sand, big rock).
$S^5$ Mars~\cite{Zhang2022S5MarsSA} is another dataset, sparsely annotated for semantic segmentation on Mars. 
GMSRI~\cite{GMSRI} provides terrain classification dataset consists of real Mars images and a synthetic counterpart generated by Generative Adversarial Network (GAN).
Unfortunately,~\cite{GMSRI} focuses on terrain classification, and~\cite{AI4Mars} and~\cite{Zhang2022S5MarsSA} focus on semantic segmentation, thus they do not offer pixel-wise instance masks.
This limits their usage in instance-aware tasks such as identification of environmental landmarks.
To overcome the aforementioned problems, this work presents a novel \textit{open-source} robotics simulator for lunar environments.
Similarly to \cite{ViperGazeboSim, orsula2022learning, FauxRanger, PUT-sim, edge} our simulator does not feature terramechanics model.
We provide basic algorithms for procedural environment generation, automatic  generation of annotated synthetic datasets, ROS1 and ROS2 example as well as sim-to-real evaluation of rock recognition to evaluate the quality of models learned in simulation. 

\section{Method}

\subsection{Terrain Generation and rendering}
In the following, we discuss how terrains are generated and managed inside our simulation.
We do not go over the terramechanics as they are not modeled accurately.
\subsubsection{In-Lab Acquisition}
Our simulation comes with different environments ready to be used.
One of them is the Lunalab~\cite{Ludivig2020BUILDINGAP}, a 6.5 meters by 10 meters lunar-analog facility located inside the University of Luxembourg.
To create a digital twin of this lab, we captured two real terrains and placed them inside a replica of the lab modeled inside Blender.
To acquire the terrains, we scanned the lab's ground section with a total station from three different viewpoints.
This resulted in 3 point clouds that were then aligned, merged, and projected onto a grid with a resolution of 1cm per pixel.
This grid was then manually cleaned to remove artifacts from the laser and interpolated with a cubic function to fill any gaps in the laser points.
In the end, this process gave us a pair of \glspl{dem} that can be loaded inside the simulator.

\subsubsection{Real Lunar Terrains}
In addition to these small-scale terrains collected in the lab, the user can choose between over 20 real lunar terrains. 
To generate these terrains, we rely on \gls{asp} and LRO images to generate 3km by 3km \glspl{dem} at 1 meter per pixel.
In practice, without using advanced processing methods such as \gls{sfs}, \glspl{dem} only offer details of about 10 meters in size.
This resolution is not sufficient to conduct realistic experiments~\cite{ViperGazeboSim}.
Thus, we augment these maps using GANs~\cite{GAN, SPADE, pix2pix} trained to perform DEM super-resolution conditioned on RGB images\footnote{\url{https://github.com/AntoineRichard/MoonSuperResolution}}.
This yields finer terrain features, of about 1 meters to 3 meters per pixel.
This method was chosen because unlike \gls{sfs}, once the networks are tuned, it is easy to use and does not require fiddling with the settings.
Yet, even at 1meters per pixel their resolution is not good enough for robotic lunar rover simulations~\cite{ViperGazeboSim}.
Hence, similarly to ~\cite{NASA-DLES, ViperGazeboSim} we provide the option to use procedural generation (see~\ref{sec:procedural}) to artificially enhance them down to 4cm per pixel.

\subsubsection{Procedural Generation}\label{sec:procedural}
The procedural generation allows us to generate unique random environments with fine-grain terrain features.
To do so, we generate small-scale craters from 10 meters to 0.5 meters in size.
To generate craters, we collected 100 half-crater profiles on SLDEM images~\cite{barker2016new}, and curated the profiles by manually removing outliers, normalizing them, smoothing them using polynomial filters, and fitting a cubic spline on them.
To create a new crater of arbitrary radius $N$ meters, we create a matrix of size $4N \times 1/R$, with $R$ the resolution of the terrain in meters per pixel.
For each cell of this matrix, we compute its distance to the center.
This distance is distorted and randomly rotated to create some variability in the crater shape.
Then, using one of the previously collected crater profiles, for each cell of the distance matrix, we query the spline function to get the elevation of the crater at that point.
In the end, this results in the generation of the \gls{dem} of a crater.
This process is used to generate random craters at run-time, allowing to generate thousands of unique-looking craters in less than a second.
To distribute craters, we are relying on hard-core Poisson distributions.
To generate large, medium, and small craters, we typically, set three different distribution densities each associated with its own radius ranges.
Furthermore, to enforce consistent behavior in between simulations, all randomizers are seeded.
This means that with the same seed, two simulation instances will generate the exact same terrains.
The user has the flexibility to adjust the parameters.
There is ample literature on crater densities on the moon~\cite{STOPAR201734} which the curious reader could refer to.
This process can be used without any modification in different scenes.
The lunalab for instance supports randomized terrain generation.
We also have procedural-only terrains such as the ``lunaryard'' familly of environments that mimic typical outdoor facilities for lunar simulations.

\subsubsection{Importing Terrains in Isaac}
With the generation of the terrains covered, let us see how they are being fed to IsaacSim.
To import terrains inside Isaac, we dynamically author the properties of meshes within the USD stage.
This is done by editing four properties of a USDGeom.Mesh, the FaceVertexIndices (how the vertices are connected), FaceVertexCounts (the number of vertices), Points (the position the vertices), and finally st (the uv coordinates for each element inside FaceVertexIndices).
This process is simplified by the fact that \glspl{dem} are regular grids, hence, computing these values is fairly straightforward.
An advantage of this method is that the terrain can be randomized without exiting the simulation.
Typically, a complete terrain randomization is achieved under 3 seconds, this includes generating the DEM, updating the visual mesh, and computing the collision mesh.
The computation time depends on the number of craters and the size of the map, but the most expensive operation is the computation of the collision mesh.
This ability to randomize the terrain can be particularly interesting when training RL agents.
Another benefit of this approach is that in a future update, its flexibility will allow the rovers to leave ruts in the terrain.

\subsection{Lunar Environment Modeling}
With the generation and management of the terrain out of the way, we can now take a look at the rest of the elements inside the simulation.

\subsubsection{Surface texture}
To texture, the surface of the terrains, we have different materials.
The first one is a modified gravel material from Polyhaven\footnote{~\url{https://polyhaven.com/}}, it is used to replicate the basalt inside the Lunalab.
For larger environments, we use sand-like materials as well as thinner gravel materials from Polyhaven and Nvidia's Base Material collection.
We are aware that none of the materials currently used in the simulation is close to the real ones, but to the best of our knowledge, this kind of material is not publicly available.

\subsubsection{Rocks}
Getting realistic rocks into the simulation is particularly challenging, the main issue is finding free-of-right quality lunar rocks.
For the digital twin of the Lunalab, we opted for photogrammetry.
We took the rocks we have in the lab, and collected about 200 pictures of each, these were then processed inside Reality Capture~\cite{RealityCapture} which generated high-poly meshes and textures.
These meshes were then imported into Blender where they were cleaned up, scaled down, and decimated to reach about 40k polygons.
Then, using the original high-resolution meshes, we baked the fine details into a normal map and exported them in the USD format.

In addition to these rocks, we also converted all the Apollo rocks from Astromaterials 3D~\cite{Astromaterials} to the USD format.
These rocks are the result of the 3D reconstruction of real lunar rocks brought back to Earth during the Apollo mission.

\subsubsection{Procedural asset management}
To manage assets into the scene, we rely on instancers.
An instancer keeps a cache of potentially usable assets and allows scattering millions of their copies onto ``points''.
Their main advantage lies in the fact that even if an asset is replicated thousands of times in the scene, only a single copy is stored in memory, making it extremely efficient to render large-scale environments.
To place these assets, we have built a Python library that can distribute points onto geometric primitives, and images.
It features, multiple distributions such as Mattern, or Thomas point processes that can be useful to distribute trees, but also simpler processes such as Poisson, Normal, or Uniform.
When creating their own environments, users will be able to choose and parametrize these settings.
In practice, we rely on hardcore Poisson point processes to place the rocks in the lunalab, and Thomas point processes for other environments.
Using the DEM elevation, and knowing their resolution, based on their xy coordinates the rocks can be placed at the right elevation.
Similarly to the terrains, the procedural placement of assets is seeded, so that the simulator yields the exact same rock position between two simulation runs, a must-have for reproducibility.

\subsubsection{Robot modelling}
\begin{figure}[t]
    \centering
    \begin{minipage}[t]{.50\linewidth}
        \centering
        \includegraphics[width=1.0\linewidth,clip]{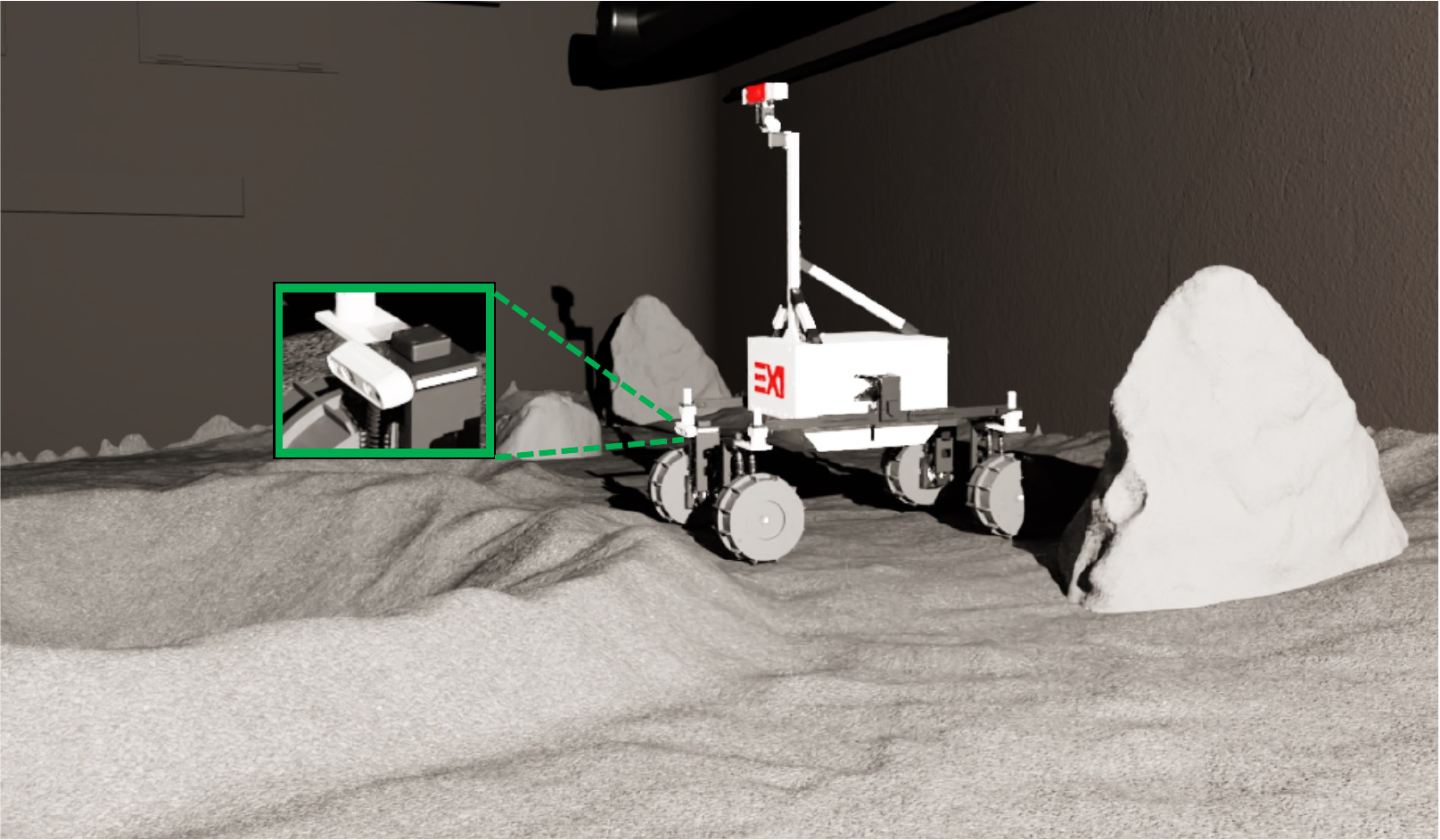}
        {\footnotesize (a) EX1 rover.}
    \end{minipage}
    \begin{minipage}[t]{.47\linewidth}
        \centering
        \includegraphics[width=1.0\linewidth,clip]{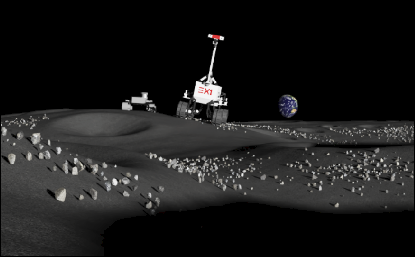}
        {\footnotesize (b) Leo and Ex1 rovers.}
    \end{minipage}
    \caption{Left: One of the robot, EX1 delivered with the simulator demonstrating a Lunalab environment. 
    Right: EX1 and a Leo rover in a larger scale environment (the real Lunar terrain).}
    \label{fig:ex1-desc}
    \vspace{-5mm}
\end{figure}

To simulate the kinematics of the rovers as well as their sensor playloads for vision tasks, we must create realistic models of rovers and sensors.
Our simulator comes with different robot models with onboard sensors ready to use: Leo rover~\cite{leo-rover} and EX1 rover~\cite{ex1}.
To simulate onboard sensor such as stereoscopic camera, appropriate camera parameters and attachment positions must be given.
As a target camera model, we chose to use Intel RealSense D435~\cite{realsense} because of its wide usage in the field of robotics.
Parameters for the simulated camera are determined from the actual sensor specifications.
As for Leo rover, the camera is attached in front of base link.
As for EX1 rover, the camera is attached in front of the right rocker arm link (0.25m from the ground) as shown in Fig~\ref{fig:ex1-desc}.

\subsubsection{ROS bindings}
Our environments and robots are completely integrated with both ROS1 and ROS2. For the robot, we use the sensors provided by default inside Isaac: RGBD cameras, 2D and 3D lidars, IMUs, Tfs, and joint-states. To control the the robots we equipped them with differential controllers that can be commanded using a Twist message. We also provide ROS topics to spawn robots, and the simulation supports multi-robots.
As for the environments, they come with a range of topics allowing to change the terrain texture, the position of lights sources, their intensity, randomize the position of the rocks, swap the terrain and more.
Our code is built such that ROS wraps around the API that controls the environment, this enables to easily add new functions accessible through ROS.


\section{Experiment}

One of the most important requirement for this kind of simulator is for the gap between simulation and the real-world to be small. 
This is crucial for perception tasks since models trained using simulation data can fail to generalize toreal world without accurate modeling~\cite{DR1}.
To evaluate if our simulator is able to perform sim-to-real transfers, we train a neural network on different dataset, synthetic and real, under different conditions.
To this end, we leverage a rock segmentation task, and see how different simulation settings impact the performance of the network when applied on real-data.

\subsection{Data Acquisition}
\subsubsection{Real-world data}
We begin with the  generation of real-world dataset.
To this effect, we recorded images inside the real Lunalab~\cite{Ludivig2020BUILDINGAP}, with a RealSense D455 camera at 5Hz saving only the RGB data.
We collected three rosbags with different light source positions: high, middle, and low\footnote{All of the annotated real Lunalab data are available at \url{https://universe.roboflow.com/srlresearch/d435_25degree_hd_6fps_nano_full_high}, \url{https://universe.roboflow.com/srlresearch/d435_25degree_hd_6fps_nano_full_mid}, \url{https://universe.roboflow.com/srlresearch/d435_25degree_hd_6fps_nano_full_low}.}.
Out of these rosbags, we kept one frame every 10 consecutive frames, which resulted in about 450 images for each.
Then, we manually annotated rocks inside each image (see Fig.~\ref{fig:sim-real-data-ex}(a) for a sample image).
We used the data with high and low light positions in the training set (955 images), and the  data with low light position in the test set (439 images).
Furthermore, we also prepare small set of test images (74 images) from the Apollo 17 mission. library~\cite{Appolo17-data}.
In those images, only rocks roughly the size of the ones in Lunalab (0.5m) are annotated\footnote{\url{https://universe.roboflow.com/srlresearch/as17-137}}.

\subsubsection{Synthetic data}
\begin{figure}[t]
    \centering
    \begin{minipage}[t]{.375\linewidth}
        \centering
        \includegraphics[width=1.0\linewidth,clip]{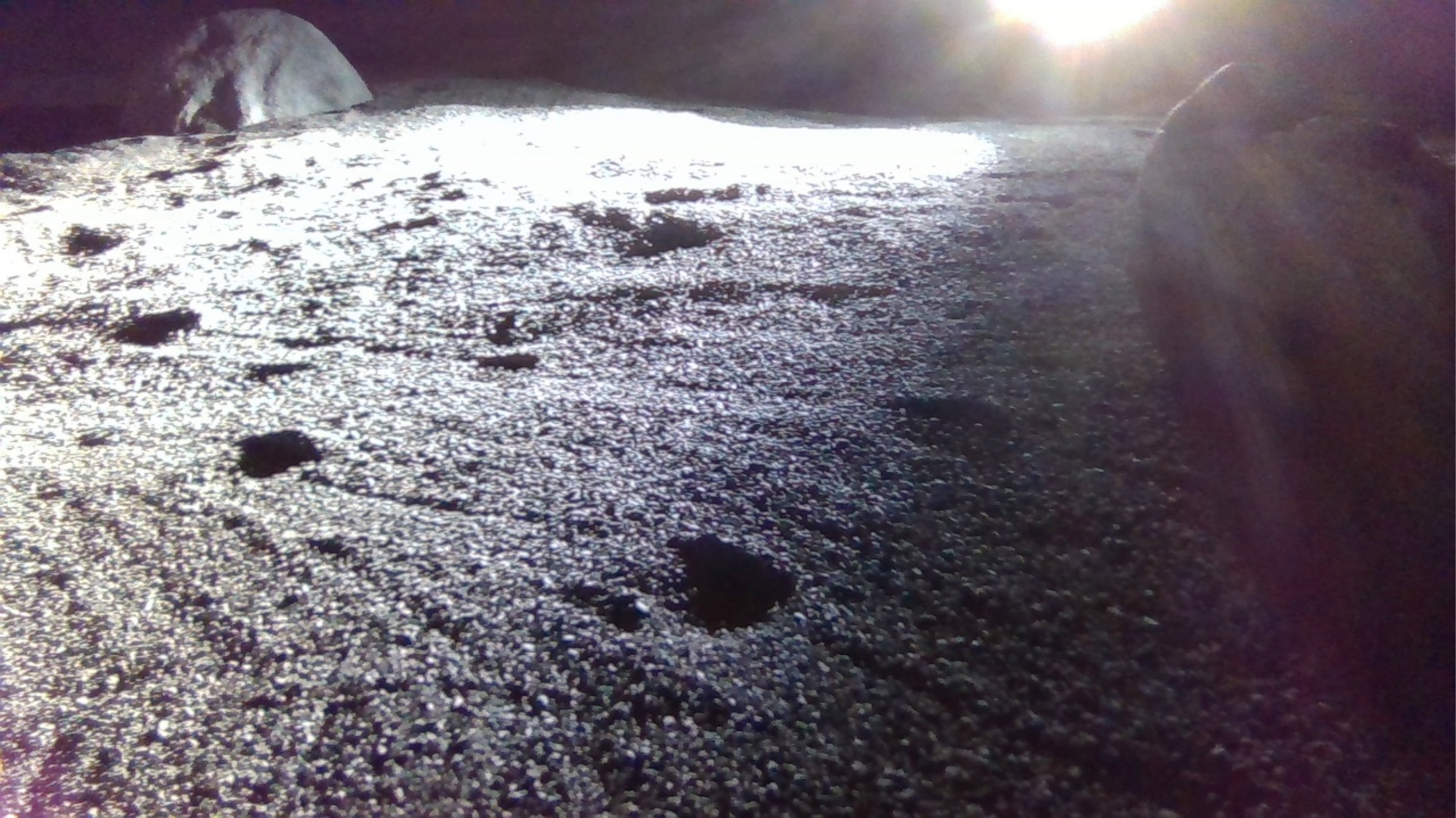}
        {\footnotesize (a) Real world.}
    \end{minipage}
    \begin{minipage}[t]{.28\linewidth}
        \centering
        \includegraphics[width=1.0\linewidth,clip]{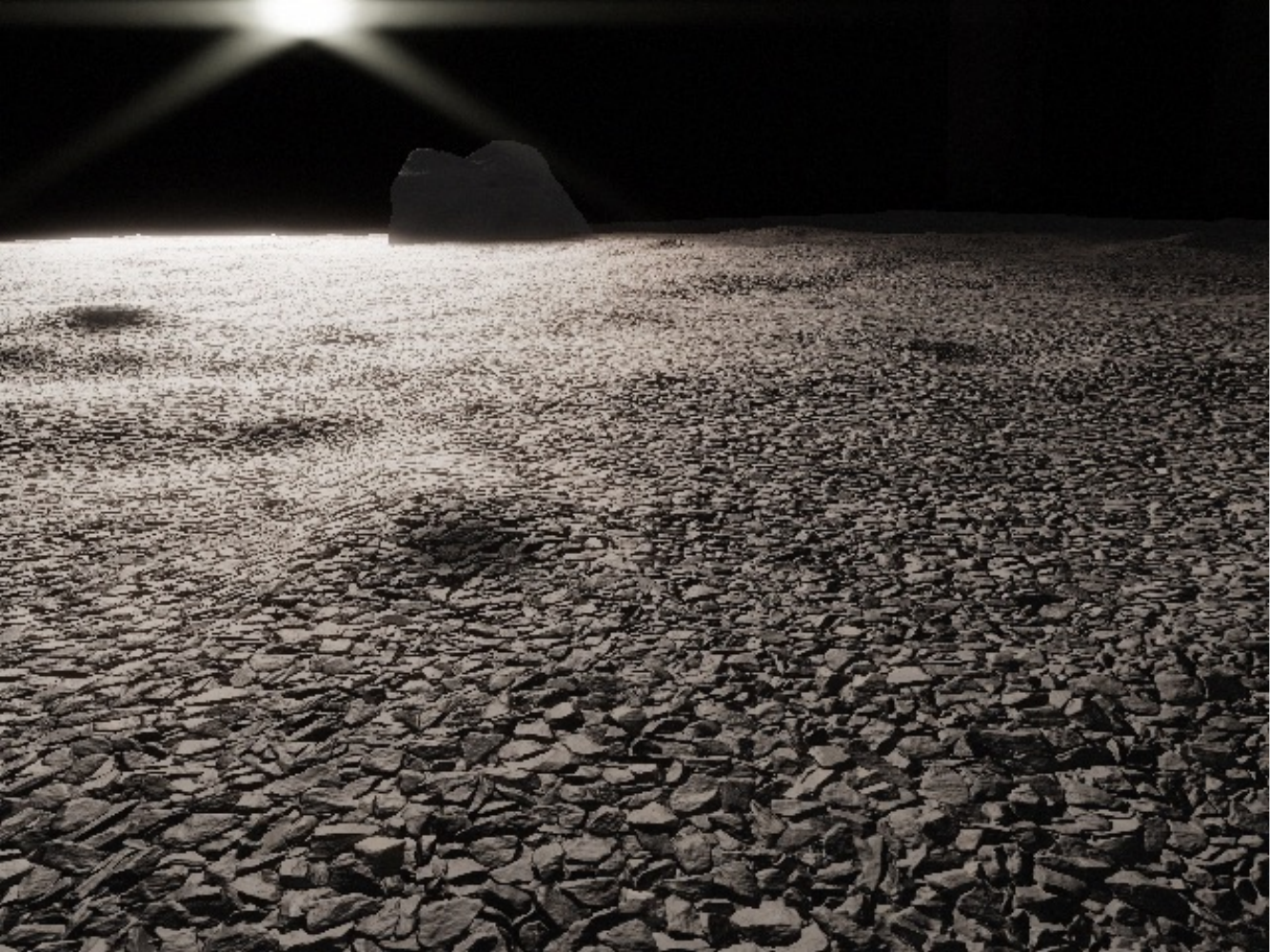}
        {\footnotesize (b) Raytrace.}
    \end{minipage}
    \begin{minipage}[t]{.28\linewidth}
        \centering
        \includegraphics[width=1.0\linewidth,clip]{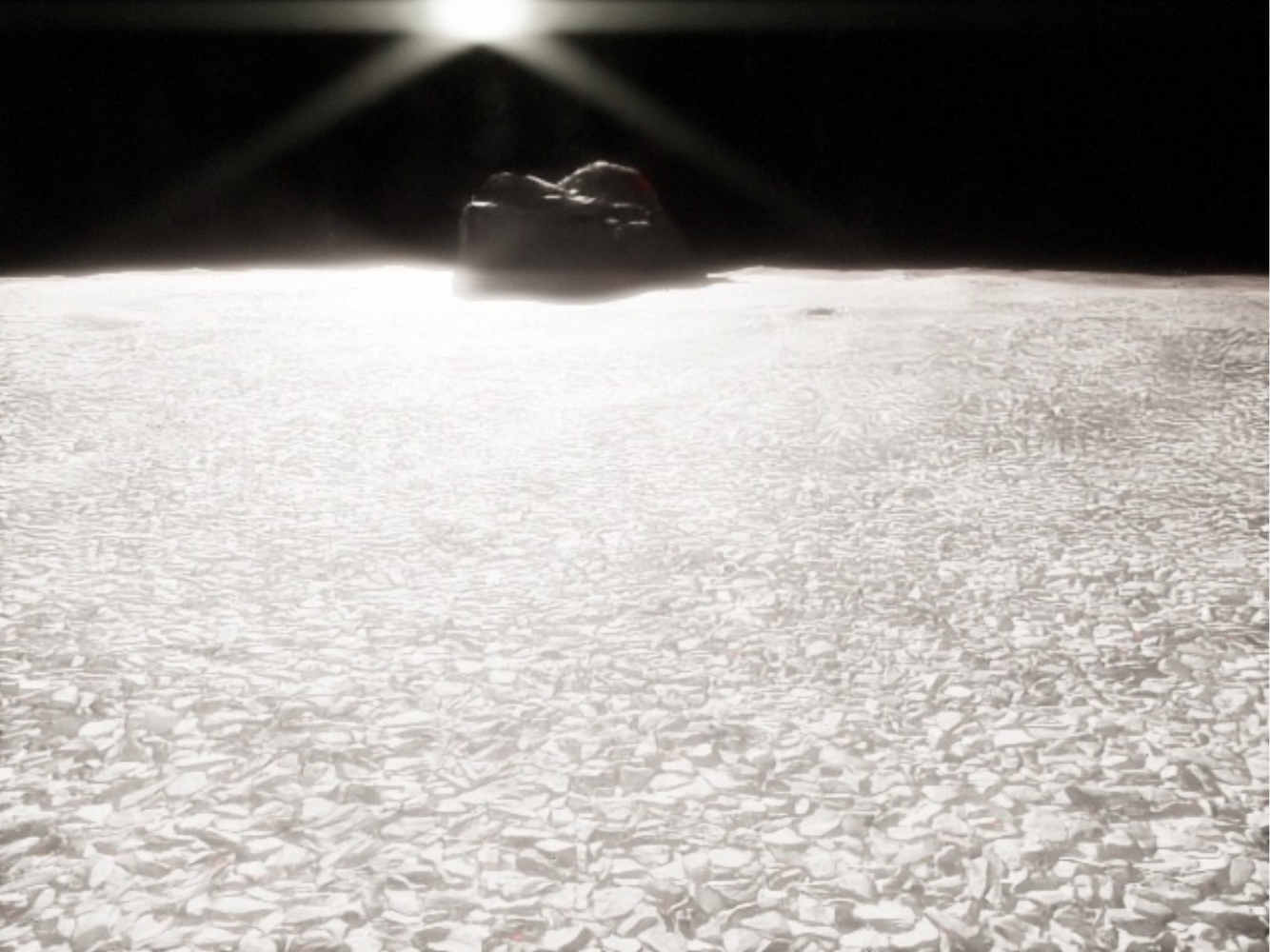}
        {\footnotesize (c) Pathtrace.}
    \end{minipage}
    \caption{Example images from both the real and simulated lunalab.}
    \label{fig:sim-real-data-ex}
    \vspace{-5mm}
\end{figure}

For the Synthetic data,  we record it inside the digital twin of the Lunalab.
To generate synthetic data, we leverage the automatic labeling from Omniverse's Replicator, a collection of tools to record synthetic data with ground-truth annotation (e.g. bounding box and segmentation mask).
To evaluate the importance of the different render modes offered inside Isaac, we render synthetic data with both RTX - Real-Time, also known as  raytracing~\cite{ray-trace} (14,099 images) and RTX - interactive, or pathtracing~\cite{path-trace} (14,099images)\footnote{This data will be released at a later time.}.
Because raytracing employs various shading approximations and optimizations to achieve both image fidelity and high frame rates, it is less accurate than pathtracing when it comes to render light.
An examples of synthetic data are shown in Fig.~\ref{fig:sim-real-data-ex}(b) and (c).


\subsubsection{Domain randomization}
During the data acquisition, we apply domain randomization~\cite{DR1, DR2} to create a wide variety of images. In practice, this is done by randomizing the rock types and as well as their locations, the light source intensity and location, and the shape of the terrain.
For the randomization of rocks, their positions in the xy coordinate plane are sampled from a uniform distribution, and their elevations are determined using the \gls{dem} matching the loaded terrain mesh.
Finally, to move the camera around the lab, four different rover trajectories are prerecorded via manual operation and played back during the synthetic data acquisition.

\section{Results and Analysis}
To evaluate the gap between our simulators and the real world, we investigate the performance of a neural network on different types of datasets both from our simulator and the real world.
First, we investigate the impact of the render mode~(a). Second, we explore the use of this synthetic data as pretraining for the real data~(b).
Finally, we check how models trained in the lunalab data fair on real lunar images~(c).

\subsection{Instance segmentation of rocks}
For our experiments, we leverage the recently released yolov8~\cite{yolov8} object detection model and train it on the datasets presented in Table~\ref{tab:data-ablation}. It contains the type of dataset used, the model training strategies, and the purpose of ablation (a,b,c).
For all the datasets, \textit{yolov8s-seg}, the second smallest among yolov8 instance segmentation model, is trained for 50 epochs during pre-training and 20 epochs during finetuning.
Results on both real-world Lunalab data and Apollo 17 data are listed on the right side of Table~\ref{tab:data-ablation}. 

\begin{table*}[t]
     \centering
     \caption{The results of the ablation study for the different datasets and training strategies. For the renderer, raytrace and pathtrace are used. For training strategies, we compare training with (i) only synthetic data, (ii) only real-world data, and (iii) synthetic data for pre-training and real-world data for finetuning. The last two columns show the average precision (AP and AP50 based on Microsoft COCO benchmark~\cite{MSCOCO}) of each model on the real-world test set. Best scores are highlighted in bold.}
     \begin{tabular}{l|c|cc|c|cc}
        \hline
        \multicolumn{1}{c|}{ \multirow{2}{*}{Viewpoint} } & \multicolumn{1}{c|}{ \multirow{2}{*}{Case} } & \multicolumn{2}{c|}{Training Dataset and Conditions} & \multicolumn{1}{c|}{ \multirow{2}{*}{Test Dataset} } & \multicolumn{2}{c}{Real Test Set} \\
                             \empty & \empty & Training Strategy & Renderer & \empty & AP & AP50 \\
        \hline
        \multirow{3}{11em}{(a) Impact of texture photorealism } & (a)-1 & \multicolumn{1}{l}{(ii) Training w/ only real Lunalab data} & - & \multicolumn{1}{l|}{Real Lunalab data} &  \textbf{64.7} & \textbf{84.2} \\
        \empty & (a)-2 & \multicolumn{1}{l}{(i) Training w/ only synthetic Lunalab data} & \textit{raytrace} & \multicolumn{1}{l|}{Real Lunalab data}  & 25.9 & 34.2 \\
        \empty & (a)-3 & \multicolumn{1}{l}{(i) Training w/ only synthetic Lunalab data} & \textit{pathtrace} & \multicolumn{1}{l|}{Real Lunalab data} &  59.3 & 73.6 \\
        \hline
         \multirow{3}{11em}{(b) Contribution of synthetic data pre-training} & (b)-1 & \multicolumn{1}{l}{(ii) Training w/ only real Lunalab data} & - & \multicolumn{1}{l|}{Real Lunalab data} &  64.7 & 84.2 \\
        \empty & (b)-2 & \multicolumn{1}{l}{(iii) Pre-training w/ synth. + finetuning w/ real Lunalab data} & \textit{raytrace} & \multicolumn{1}{l|}{Real Lunalab data} &  78.5 & \textbf{93.7} \\
        \empty & (b)-3 & \multicolumn{1}{l}{(iii) Pre-training w/ synth. + finetuning w/ real Lunalab data} & \textit{pathtrace} & \multicolumn{1}{l|}{Real Lunalab data} &  \textbf{79.2} & \textbf{93.7} \\
        \hline
        \multirow{3}{11em}{(c) Applicability to the actual Moon } & (c)-1 & \multicolumn{1}{l}{(ii) Training w/ only real Lunalab data} & - & \multicolumn{1}{l|}{Apollo Lunar data} &  2.8 & 4.7 \\
         & (c)-2 & \multicolumn{1}{l}{(i) Training w/ only synthetic Lunalab data} & \textit{raytrace} & \multicolumn{1}{l|}{Apollo Lunar data} &  8.0 & 13.0\\
         & (c)-3 & \multicolumn{1}{l}{(i) Training w/ only synthetic Lunalab data} & \textit{pathtrace} & \multicolumn{1}{l|}{Apollo Lunar data} & \textbf{11.4} & \textbf{14.6} \\
        \hline
     \end{tabular}
     \label{tab:data-ablation}
     \vspace{-3mm}
 \end{table*}


\subsection{Analysis and discussion}
\subsubsection{Renderer ablation}
To investigate the importance of the renderer (viewpoint (a)), we compare the model performance among raytrace rendering, pathtrace rendering, and the real-world data (referred to as baseline).
On the second row of Table~\ref{tab:data-ablation}, we can see that the model trained with raytraced images yields the lowest average precision score. However, training with pathtraced data produces an almost 40\% higher average precision score.
We believe this is related to the ability of pathtracing to produce pitch-black shadow and intense surface reflections. These are expected characteristics seen in real-world images, hence the model trained on this data shows better transferability.
From these results,  we can see that, as expected, training with pathtraced data, the most photorealistic render, produces the best sim-to-real transfer. 

\subsubsection{Training strategy ablation}
\begin{figure}[t]
    \centering
    \begin{minipage}[t]{.235\linewidth}
      \centering
      \includegraphics[width=1.0\linewidth,clip]{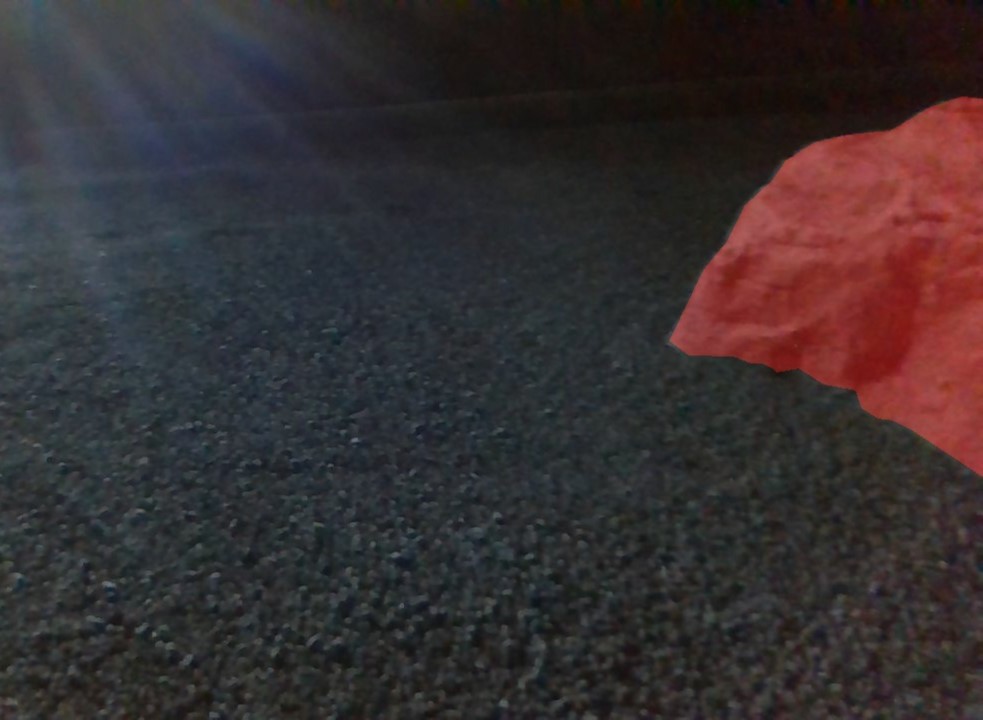}\vspace{-1mm}
      {\footnotesize Ground truth.}
    \end{minipage}
    \begin{minipage}[t]{.235\linewidth}
      \centering
      \includegraphics[width=1.0\linewidth,clip]{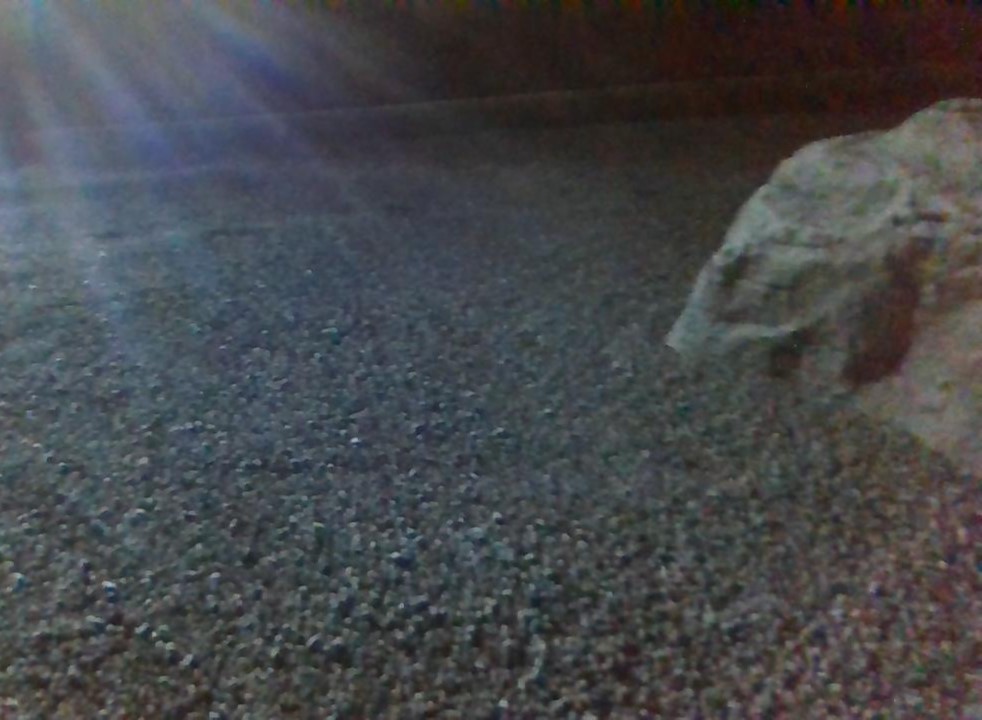}\vspace{-1mm}
      {\footnotesize (b)-1 result.}
    \end{minipage} 
    \vspace{1mm}
    \begin{minipage}[t]{.235\linewidth}
      \centering
      \includegraphics[width=1.0\linewidth,clip]{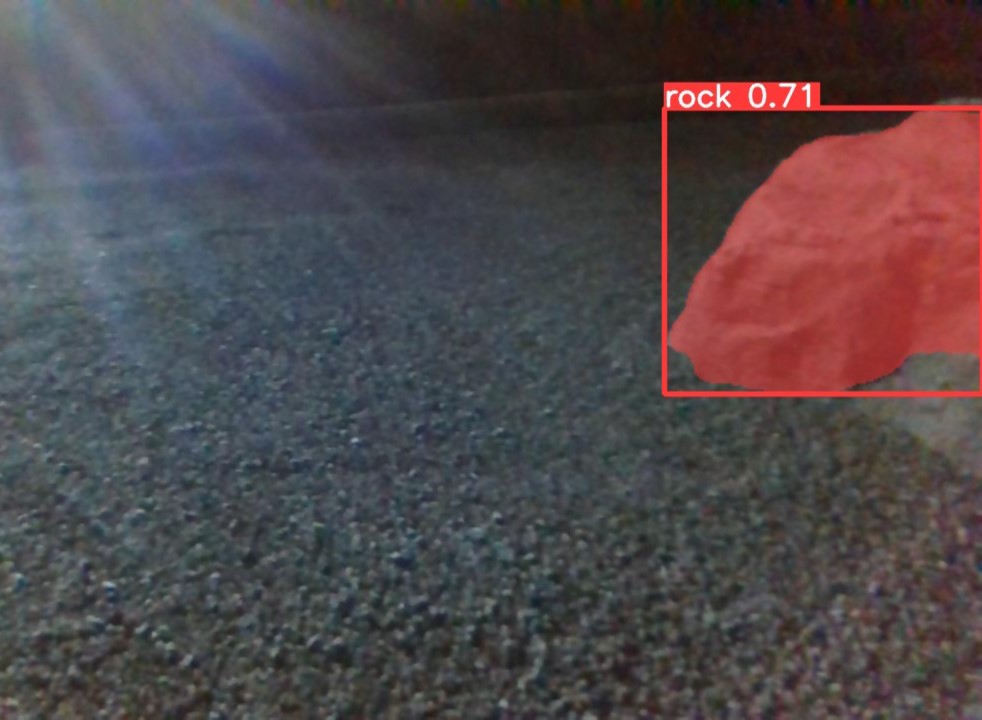}\vspace{-1mm}
      {\footnotesize (b)-2 result.}
    \end{minipage}
    \begin{minipage}[t]{.235\linewidth}
      \centering
      \includegraphics[width=1.0\linewidth,clip]{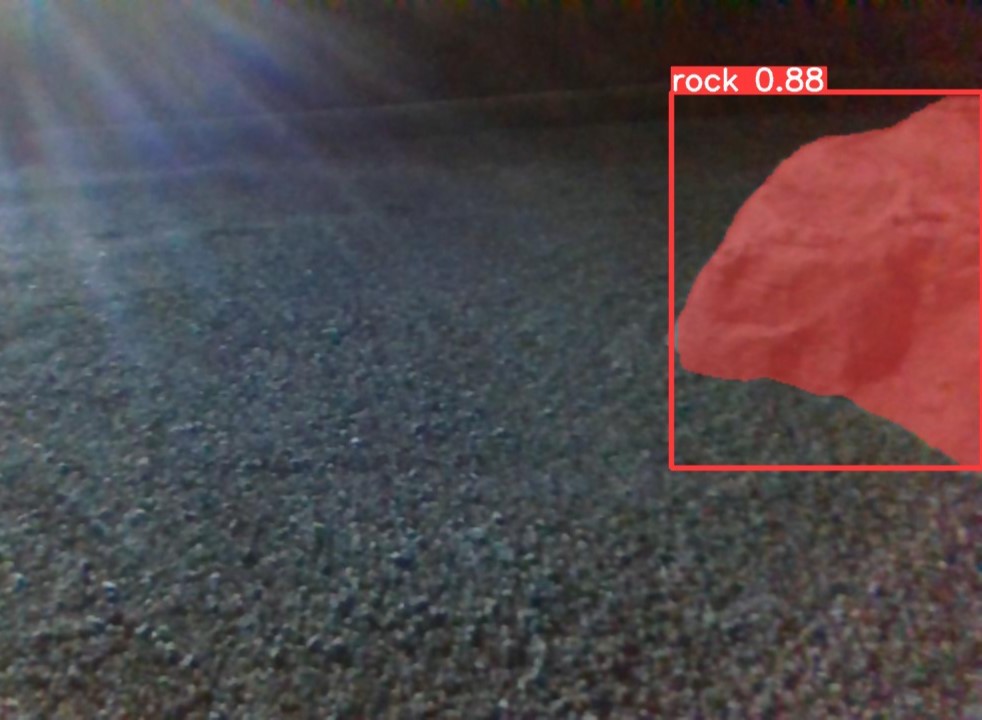}\vspace{-1mm}
      {\footnotesize (b)-3 result.}
    \end{minipage} \\ \vspace{1mm}
    {\small Segmentation result of viewpoint (b) in Table~\ref{tab:data-ablation}.}\\
    \vspace{4mm}
    \centering
    \begin{minipage}[t]{.235\linewidth}
      \centering
      \includegraphics[width=1.0\linewidth,clip]{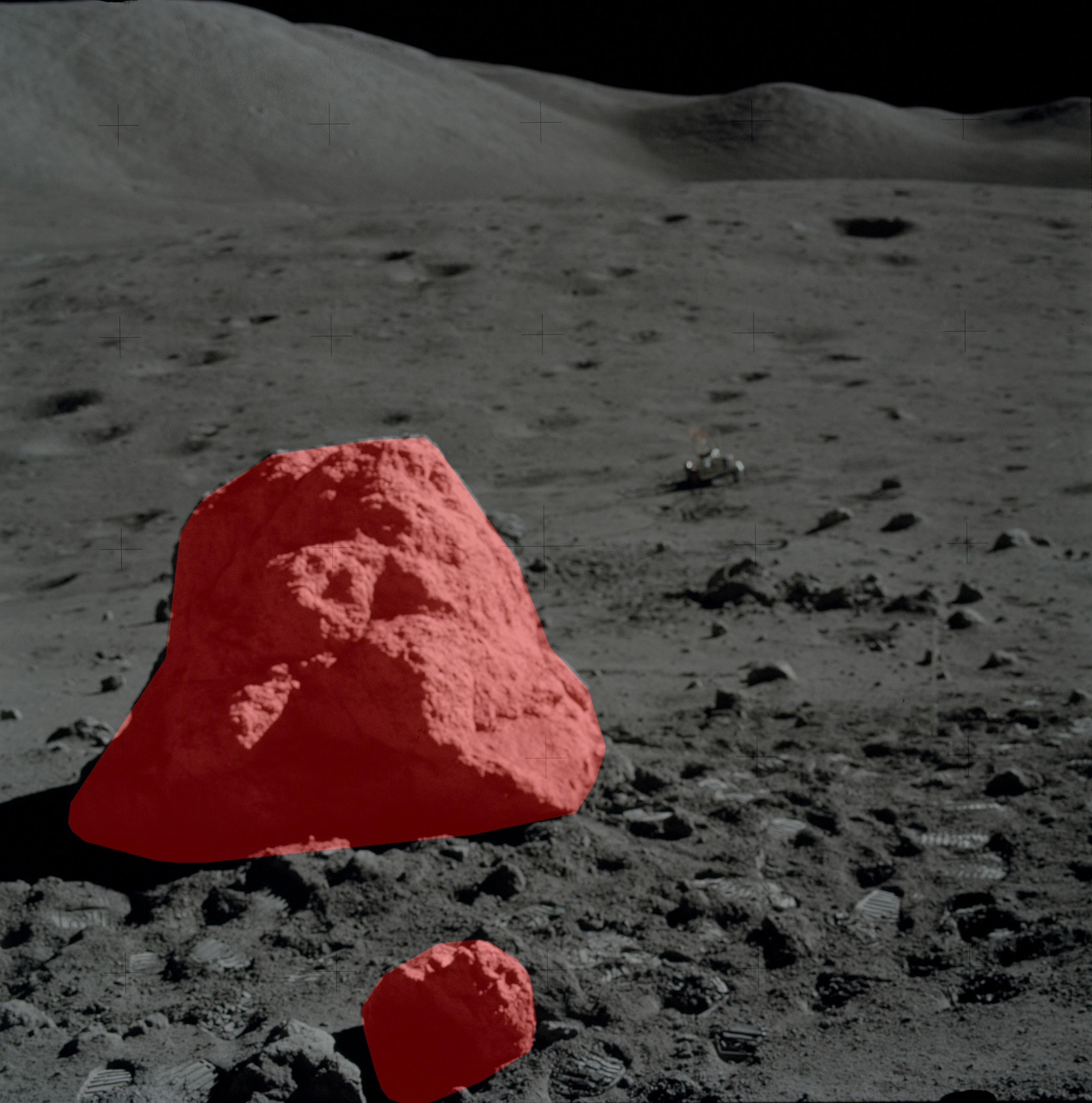}\vspace{-1mm}
      {\footnotesize Ground truth.}
    \end{minipage}
    \begin{minipage}[t]{.235\linewidth}
      \centering
      \includegraphics[width=1.0\linewidth,clip]{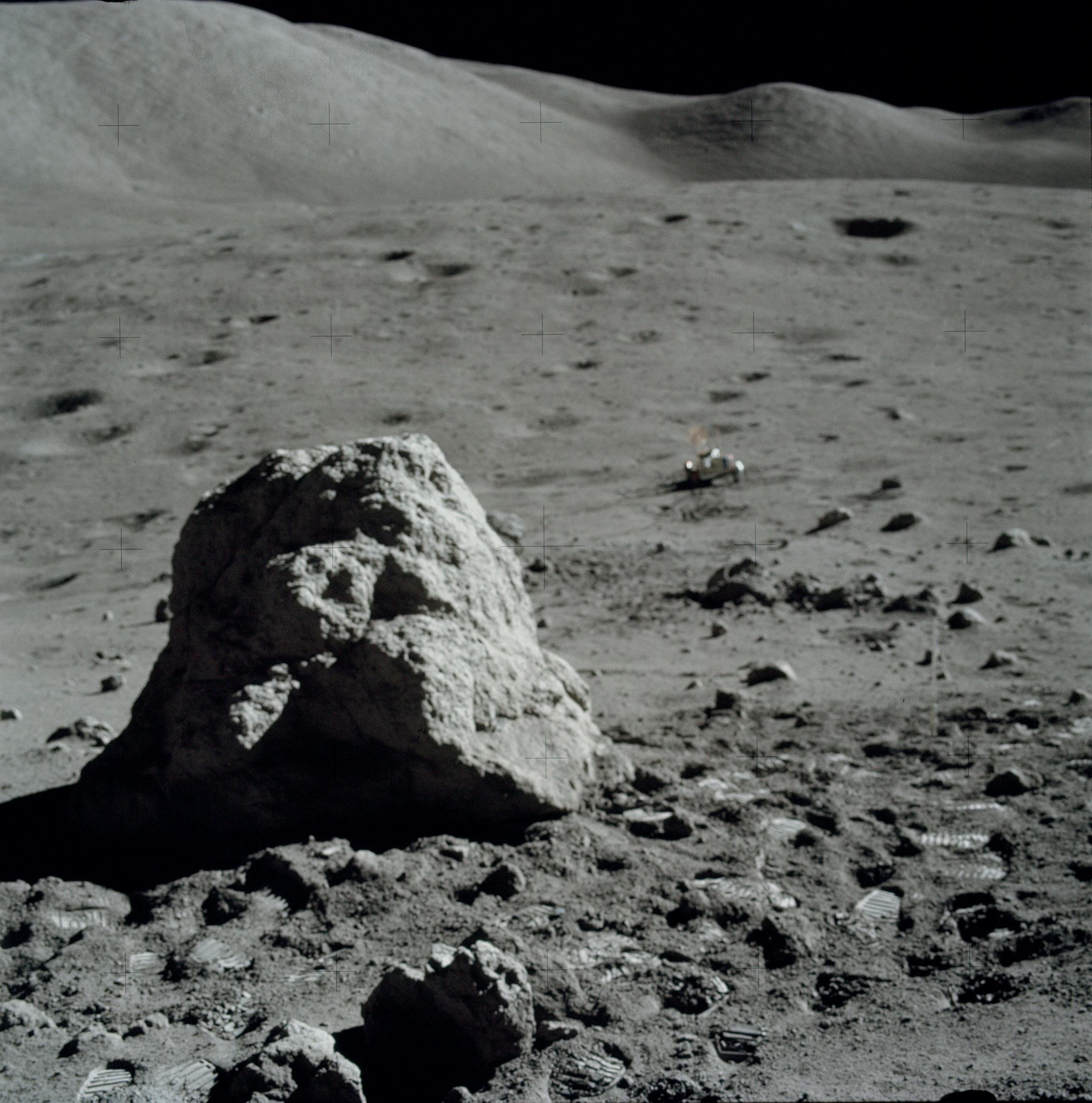}\vspace{-1mm}
      {\footnotesize (c)-1 result.}
    \end{minipage}
    \begin{minipage}[t]{.235\linewidth}
      \centering
      \includegraphics[width=1.0\linewidth,clip]{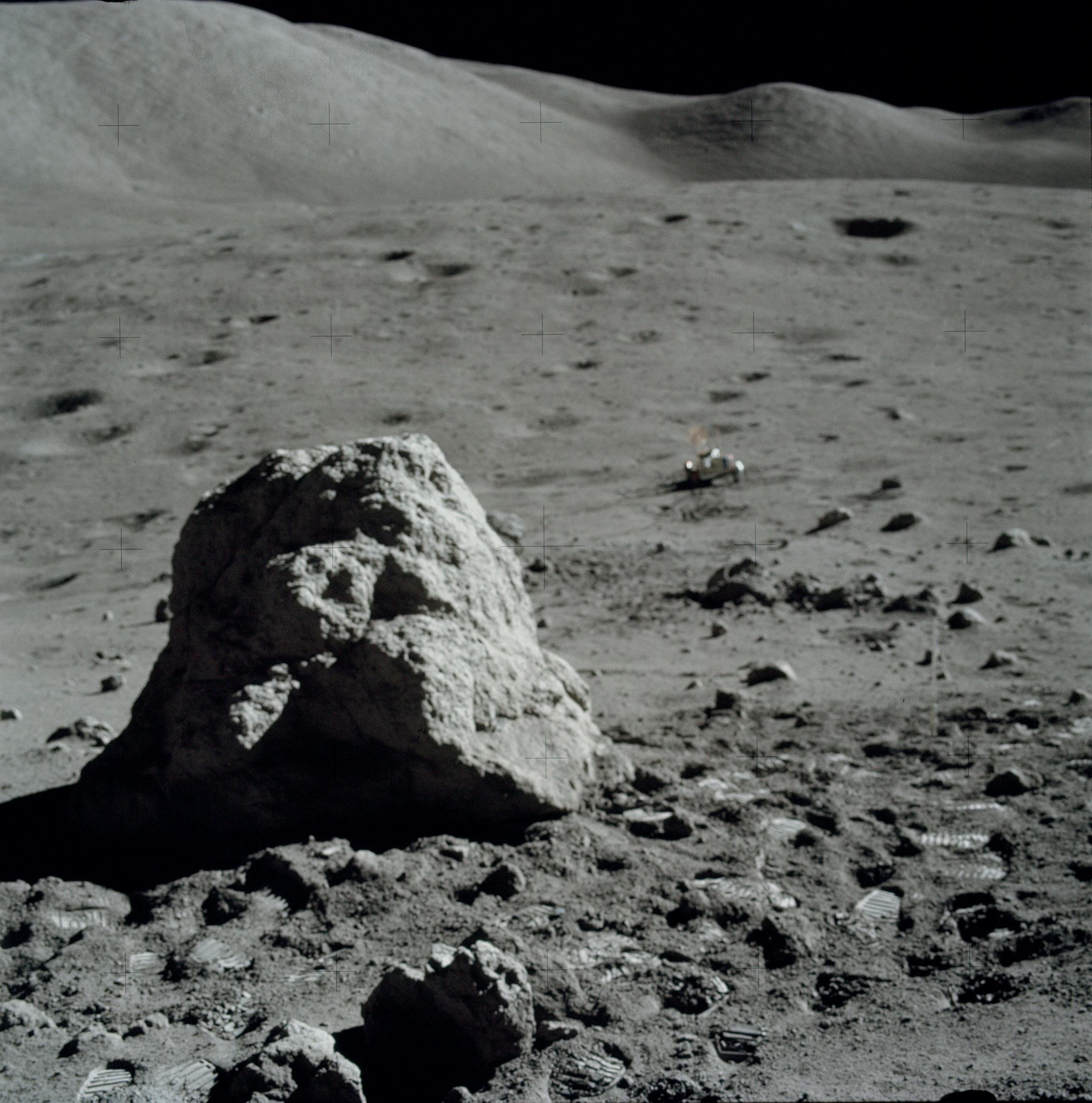}\vspace{-1mm}
      {\footnotesize (c)-2 result.}
    \end{minipage}
    \begin{minipage}[t]{.235\linewidth}
      \centering
      \includegraphics[width=1.0\linewidth,clip]{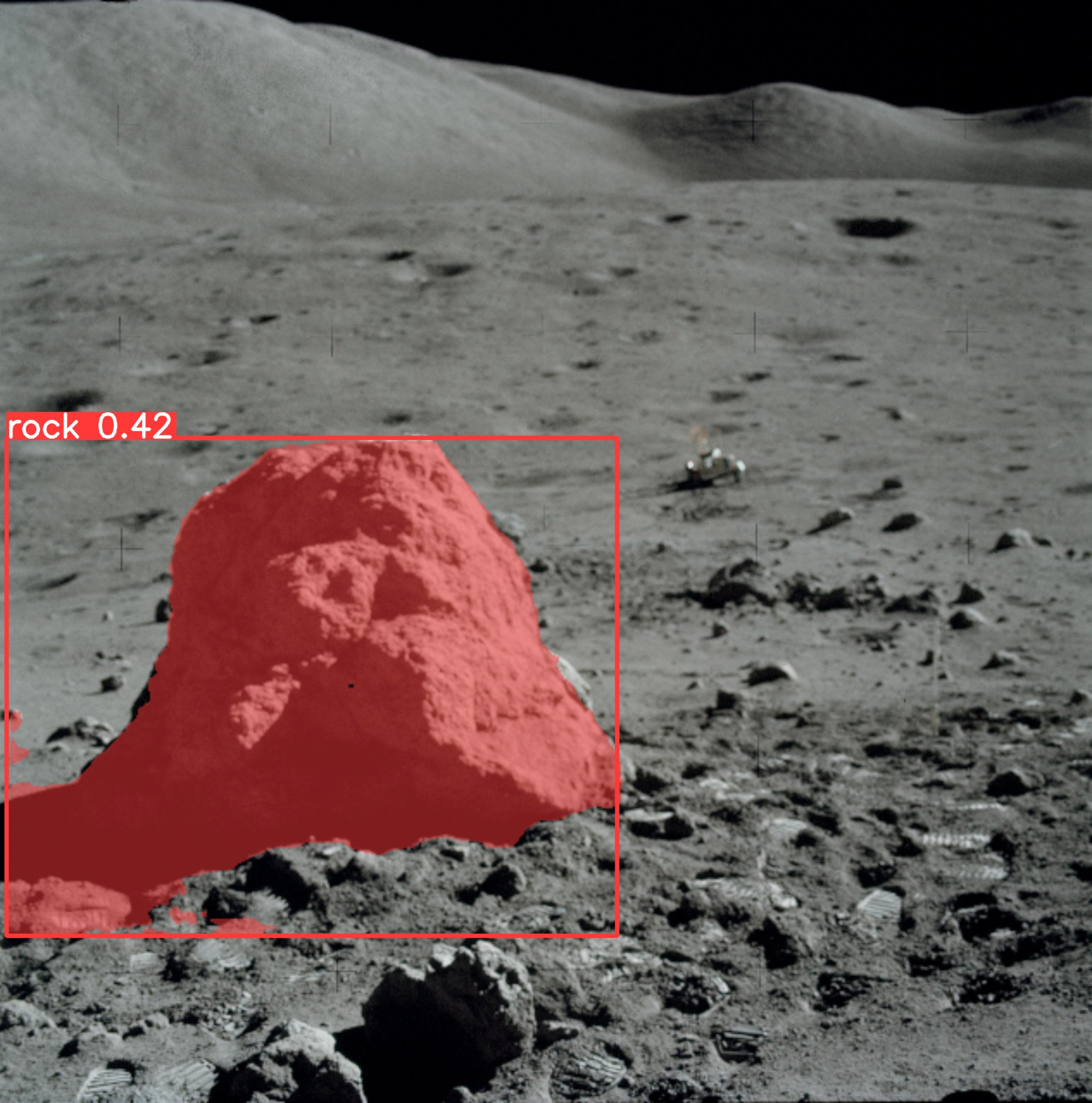}\vspace{-1mm}
      {\footnotesize (c)-3 result.}
    \end{minipage} \\ \vspace{2mm}
    {\small Segmentation result of viewpoint (c) in Table~\ref{tab:data-ablation}.}\\
    \caption{Qualitative results of instance segmentation on Lunalab test set (upper row) and Apollo mission images (lower row).}
    \label{fig:qualitative-result}
    \vspace{-5mm}
\end{figure}

Additionally, we want to find out the benefits of using synthetic data for pre-training (viewpoint (b)).
To investigate this, the model is trained with three different strategies and tested on the same test set (Lunalab recording with mid-light elevation).
As demonstrated in table~\ref{tab:data-ablation}, by pre-training on synthetic data (see fifth and sixth column), the average precision score increases by 13.8\% and 14.5\% respectively.
Also, qualitatively, while the baseline model fails to recognize rocks under challenging lens flare, our finetuned model predicts the rock's mask successfully as shown in Fig.~\ref{fig:qualitative-result} (upper row).
This suggests that our synthetic data provides good initializations weights for the model to learn from.
This finding supports the idea that when large real-world datasets are not available, a synthetic dataset can be used to provide the model with a good starting point.

\subsection{Applicability to real Lunar data}
Finally, we investigate the capability of the model's transferability to the real moon data (viewpoint (c)).
To recognize rocks in real lunar images, the model should be able to account for (1) the environmental difference, (2) the camera difference, and (3) the rock size difference.
In other words, the feature extractor of the model should learn a good representation of rocks from synthetic data to recognize rocks in a completely out-of-the-distribution test set.
To investigate the above, all the models trained (see (c) in table~\ref{tab:data-ablation}) are tested on the Apollo 17 images.
Quantitatively, the mean average precision score for all the models is low, suggesting there are strong difficulties to generalize to the real moon data.
Particularly, the baseline model produces the lowest score.
A possible reason is that since only four types/shapes of rocks are included in Lunalab when recording real and synthetic data, the model is thus biased to detect rocks in specific shapes and sizes.
This bias is alleviated in the model trained on synthetic data since the diversity in camera perspectives and the location of rocks can result in variations in the perceived size of rocks within the data. Though the lack of diversity in rock shape is still limiting the transferability of the model.
As Fig.~\ref{fig:qualitative-result} (lower row) suggests, only the model trained on synthetic pathtraced data can detect rocks successfully, while all the other models completely fail to recognize rocks.
A coherent conclusion from these evaluations would be that both diversity in rock geometry and textures are crucial for the sim-to-real transfer.

\section{Discussion}



In summary, we found that by preparing quality assets (terrain, rocks, lab props), lighting and textures, neural networks can show optimum transfer from our simulation to the real world.
In the investigation of viewpoint(a), we showed that a neural network trained on pathtraced synthetic data achieves performances close to the model trained solely on real-world data.
Furthermore, in the study of viewpoint(b), we demonstrated that neural networks trained on pathtraced synthetic data and finetuned with real-world data performs better than the model trained with only real-world data.
Although there are still many sources of sim-to-real gap that are not filled (e.g. proper lens flare, saturation and noise from sensors, and wheel ruts left by rovers), the simulation shows promising transfer capabilities, supporting the effectiveness of our simulator for vision-based tasks.

At this moment, unfortunately, we lack libraries of quality assets and textures of "real" lunar environment. Without quality assets, sim-to-real is hard to achieve as demonstrated in viewpoint(c). 
Though this limitation is global, and the field of space robotics in general would benefit from a free-of-right databank of assets and textures.
In future work, rocks with various geometry and textures will be integrated, terrain textures will be created from regolith simulant, and the procedural distribution of rocks will be further improved by using more complex probabilistic models or the actual distribution of rocks as seen in LRO images.
 

\section{Conclusions}
In this paper, we presented a novel simulator for robotics in lunar environments.
We introduced a set of open-source tools to easily simulate lunar environments, real or procedural, and perform robotic activities on them.
Along with all the implementations to generate and manipulate the environment, terrains, rocks, and ROS demonstrations are also released.
To show the realism of the generated environment, we trained a yolov8 instance segmentation network to segment rocks with different datasets and training strategies.
We found that the model trained on synthetic Lunalab data achieves performance close to the model trained on real Lunalab data, with a 5\% performance gap.
Additionally, when finetuned with real-world data, the model achieves 14\% higher average precision than the model trained only on real-world data. 
We also tested the model on Apollo 17 images, yet further works are needed to create quality asset of "real" lunar environment and close the sim-to-real gap.






\bibliography{./IEEEabrv,reference.bib}

\begin{thebibliography}{10}
\providecommand{\url}[1]{#1}
\csname url@rmstyle\endcsname
\providecommand{\newblock}{\relax}
\providecommand{\bibinfo}[2]{#2}
\providecommand\BIBentrySTDinterwordspacing{\spaceskip=0pt\relax}
\providecommand\BIBentryALTinterwordstretchfactor{4}
\providecommand\BIBentryALTinterwordspacing{\spaceskip=\fontdimen2\font plus
\BIBentryALTinterwordstretchfactor\fontdimen3\font minus
  \fontdimen4\font\relax}
\providecommand\BIBforeignlanguage[2]{{%
\expandafter\ifx\csname l@#1\endcsname\relax
\typeout{** WARNING: IEEEtran.bst: No hyphenation pattern has been}%
\typeout{** loaded for the language `#1'. Using the pattern for}%
\typeout{** the default language instead.}%
\else
\language=\csname l@#1\endcsname
\fi
#2}}

\bibitem{NASA-DLES}
E.~Z. Crues, S.~J. Lawrence, P.~Bielski, A.~B. Jacobs, J.~Schlueter, A.~Jagge,
  C.~Foreman, C.~Raymond, and N.~Davis, ``Digital lunar exploration sites
  (dles),'' in \emph{2022 IEEE Aerospace Conference (AERO)}, 2022, pp. 1--13.

\bibitem{DUST}
L.~Bingham, J.~Kincaid, B.~Weno, N.~Davis, E.~Paddock, and C.~Foreman,
  ``Digital lunar exploration sites unreal simulation tool (dust),'' in
  \emph{2023 IEEE Aerospace Conference}, 2023, pp. 1--12.

\bibitem{ViperGazeboSim}
M.~Allan, U.~Wong, P.~M. Furlong, A.~Rogg, S.~McMichael, T.~Welsh, I.~Chen,
  S.~Peters, B.~Gerkey, M.~Quigley, \emph{et~al.}, ``Planetary rover simulation
  for lunar exploration missions,'' in \emph{2019 IEEE Aerospace
  Conference}.\hskip 1em plus 0.5em minus 0.4em\relax IEEE, 2019, pp. 1--19.

\bibitem{Pangu}
I.~Martin, M.~Dunstan, and M.~S. Gestido, ``Planetary surface image generation
  for testing future space missions with pangu,'' in \emph{2nd RPI Space
  Imaging Workshop}.\hskip 1em plus 0.5em minus 0.4em\relax Sensing,
  Estimation, and Automation Laboratory, 2019.

\bibitem{Gazebo}
N.~Koenig and A.~Howard, ``Design and use paradigms for gazebo, an open-source
  multi-robot simulator,'' in \emph{2004 IEEE/RSJ International Conference on
  Intelligent Robots and Systems (IROS) (IEEE Cat. No.04CH37566)}, vol.~3,
  2004, pp. 2149--2154 vol.3.

\bibitem{orsula2022learning}
A.~Orsula, S.~B{\o}gh, M.~Olivares-Mendez, and C.~Martinez, ``Learning to grasp
  on the moon from 3d octree observations with deep reinforcement learning,''
  in \emph{2022 IEEE/RSJ International Conference on Intelligent Robots and
  Systems (IROS)}.\hskip 1em plus 0.5em minus 0.4em\relax IEEE, 2022, pp.
  4112--4119.

\bibitem{kilic2023multi}
C.~Kilic, C.~Tatsch, G.~Pereira, J.~Gross, \emph{et~al.}, ``Multi-robot
  cooperation for lunar in-situ resource utilization,'' \emph{Frontiers in
  Robotics and AI}, vol.~10, 2023.

\bibitem{FauxRanger}
\BIBentryALTinterwordspacing
N.~Khera, \emph{FauxRanger}, 2020. [Online]. Available:
  \url{https://dev.azure.com/nkhera/_git/FauxRanger}
\BIBentrySTDinterwordspacing

\bibitem{PUT-sim}
\BIBentryALTinterwordspacing
B.~P. Dominik~Pieczynski, \emph{LunarSim - ROS2-Connected Lunar Rover
  Simulation}, 2023. [Online]. Available:
  \url{https://github.com/PUTvision/LunarSim}
\BIBentrySTDinterwordspacing

\bibitem{edge}
\BIBentryALTinterwordspacing
NASA, ``Engineering doug graphics for exploration (edge),'' accessed:
  2023-09-15. [Online]. Available:
  \url{https://software.nasa.gov/software/MSC-24663-1}
\BIBentrySTDinterwordspacing

\bibitem{AmesSP}
\BIBentryALTinterwordspacing
Z.~Moratto, M.~Broxton, R.~A. Beyer, M.~Lundy, and K.~Husmann, ``Ames stereo
  pipeline, nasa's open source automated stereogrammetry software,'' 2010.
  [Online]. Available: \url{https://api.semanticscholar.org/CorpusID:127931576}
\BIBentrySTDinterwordspacing

\bibitem{PixarUSD}
``{Pixar} universal scene description,''
  https://openusd.org/release/index.html, accessed: 2023-09-15.

\bibitem{yolov8}
\BIBentryALTinterwordspacing
G.~Jocher, A.~Chaurasia, and J.~Qiu, ``{YOLO by Ultralytics},'' Jan. 2023.
  [Online]. Available: \url{https://github.com/ultralytics/ultralytics}
\BIBentrySTDinterwordspacing

\bibitem{Virtual-Shadow-Map}
\BIBentryALTinterwordspacing
E.~Games, \emph{Virtual Shadow Maps}. [Online]. Available:
  \url{https://docs.unrealengine.com/5.0/en-US/virtual-shadow-maps-in-unreal-engine/}
\BIBentrySTDinterwordspacing

\bibitem{metasim}
A.~Kar, A.~Prakash, M.-Y. Liu, E.~Cameracci, J.~Yuan, M.~Rusiniak, D.~Acuna,
  A.~Torralba, and S.~Fidler, ``Meta-sim: Learning to generate synthetic
  datasets,'' 2019.

\bibitem{metasim2}
J.~Devaranjan, A.~Kar, and S.~Fidler, ``Meta-sim2: Unsupervised learning of
  scene structure for synthetic data generation,'' 2020.

\bibitem{kaggle-lunar-dataset}
\BIBentryALTinterwordspacing
R.~Pessia \emph{et~al.}, 2020. [Online]. Available:
  \url{https://www.kaggle.com/datasets/romainpessia/artificial-lunar-rocky-landscape-dataset}
\BIBentrySTDinterwordspacing

\bibitem{ReSyRIS}
W.~Boerdijk \emph{et~al.}, ``Resyris - a real-synthetic rock instance
  segmentation dataset for training and benchmarking,'' in \emph{2023 IEEE
  Aerospace Conference}, 2023, pp. 1--9.

\bibitem{DLR-OAISYS}
M.~G. Müller \emph{et~al.}, ``A photorealistic terrain simulation pipeline for
  unstructured outdoor environments,'' in \emph{2021 IEEE/RSJ International
  Conference on Intelligent Robots and Systems (IROS)}, 2021, pp. 9765--9772.

\bibitem{GMSRI}
\BIBentryALTinterwordspacing
C.~Wang, Z.~Zhang, Y.~Zhang, R.~Tian, and M.~Ding, ``Gmsri: A texture-based
  martian surface rock image dataset,'' \emph{Sensors}, vol.~21, no.~16, 2021.
  [Online]. Available: \url{https://www.mdpi.com/1424-8220/21/16/5410}
\BIBentrySTDinterwordspacing

\bibitem{Zhang2022S5MarsSA}
\BIBentryALTinterwordspacing
J.~Zhang, L.~Lin, Z.~Fan, W.~Wang, and J.~Liu, ``S5mars: Self-supervised and
  semi-supervised learning for mars segmentation,'' \emph{ArXiv}, vol.
  abs/2207.01200, 2022. [Online]. Available:
  \url{https://api.semanticscholar.org/CorpusID:250264816}
\BIBentrySTDinterwordspacing

\bibitem{AI4Mars}
R.~M. Swan \emph{et~al.}, ``Ai4mars: A dataset for terrain-aware autonomous
  driving on mars,'' in \emph{2021 IEEE/CVF Conference on Computer Vision and
  Pattern Recognition Workshops (CVPRW)}, 2021, pp. 1982--1991.

\bibitem{SPOCDL}
B.~Rothrock \emph{et~al.}, ``Spoc: Deep learning-based terrain classification
  for mars rover missions,'' 2016.

\bibitem{Ludivig2020BUILDINGAP}
P.~Ludivig \emph{et~al.}, ``Building a piece of the moon: Construction of two
  indoor lunar analogue environments,'' 2020.

\bibitem{GAN}
I.~J. Goodfellow, J.~Pouget-Abadie, M.~Mirza, B.~Xu, D.~Warde-Farley, S.~Ozair,
  A.~Courville, and Y.~Bengio, ``Generative adversarial networks,'' 2014.

\bibitem{SPADE}
T.~Park, M.-Y. Liu, T.-C. Wang, and J.-Y. Zhu, ``Semantic image synthesis with
  spatially-adaptive normalization,'' 2019.

\bibitem{pix2pix}
P.~Isola, J.-Y. Zhu, T.~Zhou, and A.~A. Efros, ``Image-to-image translation
  with conditional adversarial networks,'' 2016.

\bibitem{barker2016new}
M.~Barker, E.~Mazarico, G.~Neumann, M.~Zuber, J.~Haruyama, and D.~Smith, ``A
  new lunar digital elevation model from the lunar orbiter laser altimeter and
  selene terrain camera,'' \emph{Icarus}, vol. 273, pp. 346--355, 2016.

\bibitem{STOPAR201734}
\BIBentryALTinterwordspacing
J.~D. Stopar, M.~S. Robinson, O.~S. Barnouin, A.~S. McEwen, E.~J. Speyerer,
  M.~R. Henriksen, and S.~S. Sutton, ``Relative depths of simple craters and
  the nature of the lunar regolith,'' \emph{Icarus}, vol. 298, pp. 34--48,
  2017, lunar Reconnaissance Orbiter - Part III. [Online]. Available:
  \url{https://www.sciencedirect.com/science/article/pii/S0019103517304013}
\BIBentrySTDinterwordspacing

\bibitem{RealityCapture}
\BIBentryALTinterwordspacing
E.~Games, ``Realitycapture,'' accessed: 2023-09-15. [Online]. Available:
  \url{https://www.capturingreality.com/realitycapture}
\BIBentrySTDinterwordspacing

\bibitem{Astromaterials}
E.~Blumenfeld, K.~Beaulieu, A.~Thomas, C.~Evans, R.~Zeigler, E.~Oshel,
  D.~Liddle, K.~Righter, R.~Hanna, and R.~Ketcham, ``3d virtual astromaterials
  samples collection of nasa’s apollo lunar and antarctic meteorite samples
  to be an online database to serve researchers and the public,'' in \emph{50th
  Lunar and Planetary Science Conference, held}, 2019, pp. 18--22.

\bibitem{leo-rover}
``Leo rover,'' \url{https://www.leorover.tech/the-rover}, accessed: 2023-09-13.

\bibitem{ex1}
D.~Rodríguez-Martínez, K.~Uno, K.~Sawa, M.~Uda, G.~Kudo, G.~H. Diaz,
  A.~Umemura, S.~Santra, and K.~Yoshida, ``Enabling faster locomotion of
  planetary rovers with a mechanically-hybrid suspension,'' \emph{arXiv
  preprint arXiv:2307.04494}, 2023.

\bibitem{realsense}
L.~Keselman, J.~I. Woodfill, A.~Grunnet-Jepsen, and A.~Bhowmik, ``Intel
  realsense stereoscopic depth cameras,'' 2017.

\bibitem{DR1}
J.~Tobin, R.~Fong, A.~Ray, J.~Schneider, W.~Zaremba, and P.~Abbeel, ``Domain
  randomization for transferring deep neural networks from simulation to the
  real world,'' 2017.

\bibitem{Appolo17-data}
\BIBentryALTinterwordspacing
E.~Jones, ``Apollo 17 image library,'' 2017. [Online]. Available:
  \url{https://history.nasa.gov/alsj/a17/images17.html#MagC}
\BIBentrySTDinterwordspacing

\bibitem{ray-trace}
\BIBentryALTinterwordspacing
T.~Whitted, ``An improved illumination model for shaded display,''
  \emph{Commun. ACM}, vol.~23, no.~6, p. 343–349, jun 1980. [Online].
  Available: \url{https://doi.org/10.1145/358876.358882}
\BIBentrySTDinterwordspacing

\bibitem{path-trace}
\BIBentryALTinterwordspacing
T.~Whitted, ``An improved illumination model for shaded display,''
  \emph{Commun. ACM}, vol.~23, no.~6, p. 343–349, jun 1980. [Online].
  Available: \url{https://doi.org/10.1145/358876.358882}
\BIBentrySTDinterwordspacing

\bibitem{DR2}
J.~Tremblay, A.~Prakash, D.~Acuna, M.~Brophy, V.~Jampani, C.~Anil, T.~To,
  E.~Cameracci, S.~Boochoon, and S.~Birchfield, ``Training deep networks with
  synthetic data: Bridging the reality gap by domain randomization,'' 2018.

\bibitem{MSCOCO}
T.-Y. Lin, M.~Maire, S.~Belongie, L.~Bourdev, R.~Girshick, J.~Hays, P.~Perona,
  D.~Ramanan, C.~L. Zitnick, and P.~Dollár, ``Microsoft coco: Common objects
  in context,'' 2014.

\end{thebibliography}

\end{document}